\documentclass[letterpaper, 10 pt, conference]{ieeeconf}  % Comment this line out if you need a4paper

\IEEEoverridecommandlockouts                             
\usepackage{graphics} % for pdf, bitmapped graphics files
\usepackage{epsfig} % for postscript graphics files
\usepackage{mathptmx} % assumes new font selection scheme installed
\usepackage{times} % assumes new font selection scheme installed
\usepackage{amsmath,amssymb} % define this before the line numbering.
\usepackage{color}
\usepackage{hyperref}
\usepackage{graphicx}
\usepackage{subcaption}

\title{\LARGE \bf
MonStereo: When Monocular and Stereo Meet \\ at the Tail of 3D Human Localization
}

\author{Lorenzo Bertoni, Sven Kreiss, Taylor Mordan, Alexandre Alahi \\ Visual Intelligence for Transportation (VITA) lab, EPFL, Switzerland
}
\begin{document}

\maketitle
\thispagestyle{empty}
\pagestyle{empty}

\begin{abstract}
Monocular and stereo visions are cost-effective solutions for 3D human localization in the context of self-driving cars or social robots. However, they are usually developed independently and have their respective strengths and limitations. We propose a novel unified learning framework that leverages the strengths of both monocular and stereo cues for 3D human localization. Our method jointly (i) associates humans in left-right images, (ii) deals with occluded and distant cases in stereo settings by relying on the robustness of monocular cues, and (iii) tackles the intrinsic ambiguity of monocular perspective projection by exploiting prior knowledge of the human height distribution. We specifically evaluate outliers as well as challenging instances, such as occluded and far-away pedestrians, by analyzing the entire error distribution and by estimating calibrated confidence intervals. Finally, we critically review the official KITTI 3D metrics and propose a practical 3D localization metric tailored for humans.
\end{abstract}

\section{Introduction}

Recently, human 3D localization for autonomous vehicles or social robots has been addressed with cost-effective vision-based solutions \cite{alahi2017tracking, ku2019monopsr,monoloco,wenlongPSF,wang2019pseudo}.
All the approaches strive to improve state-of-the-art results in popular metrics.  Yet these solutions do not necessarily convey trust in real-world applications, and the long tail of 3D perception opens a Pandora's box of undetected challenges. While many methods perform very well ``on average", can they still be trusted in the most challenging cases?
The long tail of 3D object localization, \textit{i.e.}, the share of instances where methods struggle the most, is crucial for safety but rarely evaluated in standard benchmarks \cite{Geiger2013Kitti}. This is especially relevant for pedestrians, undoubtedly an essential category to safeguard from vehicle or robot collisions. 

 Multi-view \cite{alahi2008object,alahi2008master, alahi2009sparsity,alahi2017unsupervised} and stereo-based \cite{wang2018glue, wenlongPSF} methods have the potential for accurate 3D human localization, as they are free from the perspective projection ambiguity, inevitable in the monocular case \cite{monoloco}. 
 Pseudo-LiDAR \cite{wang2019pseudo} drastically reduced the discrepancy between camera and LiDAR performances by converting a stereo-based dense depth map into 3D point clouds and directly applying LiDAR-based object detectors \cite{qi2018frustum,ku2018avod}.
 However, computing depth from disparity poses two main challenges. Instances can be located out of the field-of-view or be occluded in one of the two images, and an association may not be available.
 Furthermore, a small disparity error (\textit{e.g.,} a pixel shift) for far-away objects leads to unacceptable errors of several meters, as the error grows quadratically with depth \cite{gallup2008baseline}. We identify occluded and far instances as the largest share of the stereo-based long tail of predictions.
 On the contrary, monocular images are less error-prone for far instances and do not depend on accurate detections on both images. In prior work \cite{monoloco}, we achieved competitive performances in 3D human localization by exploiting the known prior distribution of human heights. However, this approach fails in the presence of children or very tall people, connecting the long tail of monocular 3D localization with the distribution of human heights.

 \begin{figure*}[t!]
  \centering
    \includegraphics[width=\linewidth, height=5.2cm]{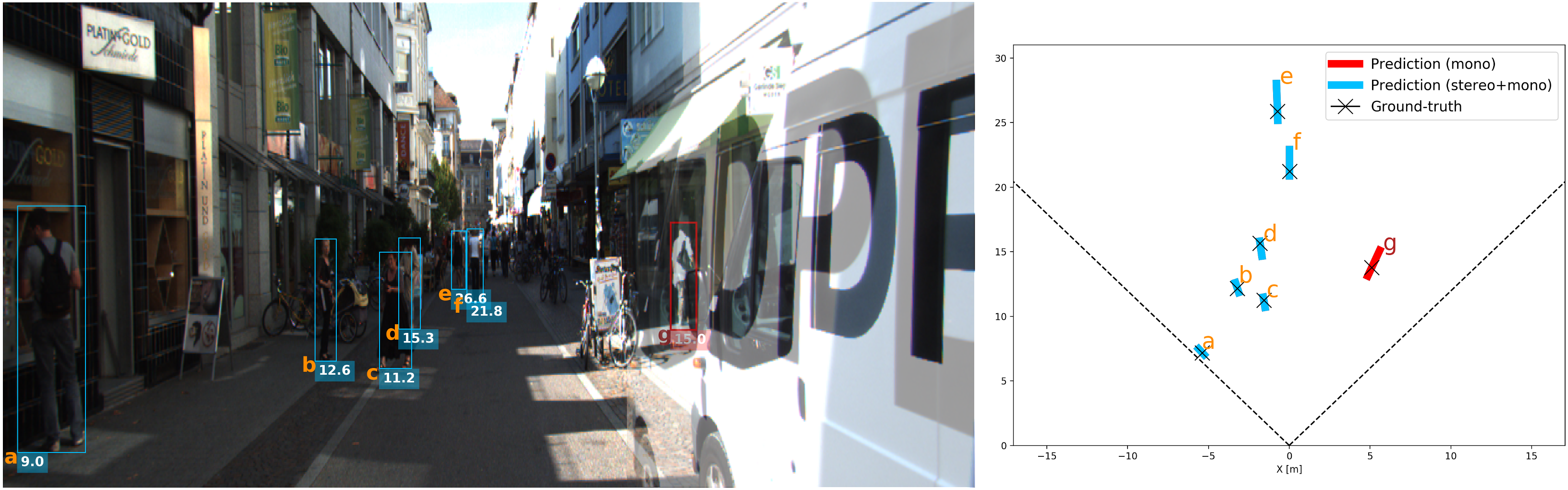}

    \caption{\small{Long tail example in the KITTI dataset~\cite{Geiger2013Kitti}. The pedestrian \textit{g} is only visible from the left camera (no stereo information available) as shown by overlapping the white van from the right image. The network classifies it as a monocular sample (red color) and outputs a larger confidence interval that reflects less accurate monocular estimates at that location. We display radial distances in meters in the frontal image and radial uncertainties in the bird-eye-view image. Only instances that match a ground-truth are shown.}}
  \label{fig:qual}
\end{figure*}

In this work, we leverage the best of both worlds, \textit{i.e}., stereo and monocular methods, in a unified learning framework tailored for pedestrian 3D localization. Our method, referred to as \textit{MonStereo},  jointly associates detections in left-right images and implicitly learns to leverage monocular and/or stereo cues. Moreover, it also learns to communicate uncertainty driven by the  cues (again without direct supervision at training time).
Our approach uses an off-the-shelf pose detector \cite{kreiss2019pifpaf, kreiss2021openpifpaf} on left-right images to obtain 2D keypoints, a low-dimensional representation of humans. 
A simple feed-forward network estimates whether each input pair is formed by the same person from left-right images and, 
concurrently, estimates the 3D location of pedestrians with their corresponding uncertainty (accounting for stereo disparity and/or monocular cues). 

The popular KITTI dataset \cite{Geiger2013Kitti} has a limited variation of instances, oversimplifying the monocular task. We address the long tail of height distribution by injecting prior knowledge from the real world. Leveraging the simplicity of manipulation of 2D keypoints, 
we create instances of people from a broader spectrum of heights. 
This conveys information about the real challenge of the task in the data domain, thus, increases the network performance and calibrates the estimated confidence intervals without the need for hand-crafted architectures. 

In summary, we propose a unified learning framework that jointly matches detections in left-right pairs of images and estimates the 3D localization of each pedestrian. We focus on the limitations of monocular and stereo visions,
referred to as the long tail challenge, by jointly exploiting stereo and monocular cues with a measure of uncertainty. We also design a data augmentation procedure to tackle the long tail of the human height distribution.  Our network achieves state-of-the-art results on 3D localization metrics and provides reliable confidence intervals even for challenging cases. Our code is available online \footnote{\url{https://github.com/vita-epfl/monoloco}}.

\section{Related Work}

\textbf{Monocular 3D Detection. }
Estimating depth from a single RGB image is an ill-posed task for non-rigid human bodies. To the best of our knowledge, few methods have explicitly tackled vulnerable road users in contrast to the large body of works related to rigid vehicles \cite{mousavian20173d,qin2019monogrnet,roddick2018orthographic,li2019gs3d,monodis}. Our recent MonoLoco \cite{monoloco} predicted confidence intervals of pedestrians to address the task ambiguity for 3D localization, while
MonoPSR \cite{ku2019monopsr} learned local shapes of objects with privileged signal at training time. Both methods, however, fail to address the long tail of 3D human localization.

\textbf{Stereo 3D Detection. }
Stereo-based 3D detectors can be grouped into \textit{instance-level} and \textit{pixel-level} depth estimators. The \textit{instance-level} approach consists of detecting instances in the image plane and comparing features of proposals in left and right frames to correctly associate objects and estimate their locations \cite{chen20153dop,li2018stereo,li2019stereo,qin2019triangulation,wenlongPSF, oc-stereo}. Among them, PSF \cite{wenlongPSF} was designed for human localization and, similarly to our method, leverages 2D keypoints to solve the association task. However, their 3D output is simply the median depth calculated from a set of disparities. 
The \textit{pixel-level} approach consists in estimating a dense disparity map for every pixel and transforming the dense map into a 3D point cloud \cite{wang2019pseudo, e2e-pl}. The pseudo-LiDAR point cloud can then be used to detect vehicles and pedestrians by applying LiDAR-based algorithms \cite{qi2018frustum,ku2018avod}. 
The underlying task of all previous methods is to compute disparity from pixels, either locally to associate and align pairs of left-right instances, or globally to find dense correspondences between pixels. Qin \textit{et al.} \cite{qin2019triangulation} have recently proposed to extend a monocular baseline to predict 3D locations of \textit{car} instances with a triangulation network. Our work goes beyond the concept of ``depth from disparity" and is not limited by the discrete nature of pixels, but can exploit together monocular and stereo cues to directly estimate a continuous depth.

\textbf{Vision-based 3D Localization Ambiguity. }
Estimating the 3D location of objects from a single RGB image is a fundamentally ill-posed problem due to the ambiguous projections to the 2D image. This is particularly true for humans due to their variation of height and non-rigid body structure. Our prior work \cite{monoloco} quantified this ambiguity as a function of the distance from the camera, assuming that the distribution of human stature follows a Gaussian distribution for male and female populations \cite{freeman1995cross}. 
The expected localization error $\hat{e}_{mono}$ due to height variations of people can be obtained by 
$\hat{e}_{mono} = C * r_{gt}$,
where the constant $C$ is modelled from the distribution of human heights and  $r_{gt}$ is the ground-truth distance.
On the other side, even if stereo methods do not suffer from the intrinsic ambiguity of perspective projection, the error grows quadratically with depth, making disparity estimation very sensitive to pixel resolution. 
The depth error $e_z$ can be expressed as a function of the disparity error $e_d$ \cite{gallup2008baseline} as 
$e_{z} \approx \frac{z^2}{b f} e_d$, 
 where $z$ is the depth, $b$ the camera baseline and $f$ the focal length.
With the goal of comparing monocular and stereo limitations,  we analyze what we call the \textit{pixel error}: the depth error due to a disparity error of one pixel. Its value depends on the characteristics of the camera and we use the camera parameters of the KITTI dataset \cite{Geiger2013Kitti}, a popular dataset for 3D object detection with stereo imaging at a resolution of $1240 \times 380$ pixels.
The results, shown in Figure  \ref{fig:spread}, highlight that the stereo depth error can become more challenging than the monocular one for humans at just over 20 meters. For example, a disparity error of 1 pixel at 40 meters corresponds to 4.5 meters of depth error. These conclusions depend on the precision of the disparity estimation and the image resolution, but highlight the importance of monocular estimation for 3D perception.

\section{Method}

The goal of our approach is to detect, associate and estimate the 3D positions of pedestrians in a pair of stereo images. We identified two main challenges for a stereo network: (i) when a person is not identified in both images, there is no disparity information and (ii) disparity estimation for faraway objects leads to poor predictions.
We propose a simple yet effective way to tackle both issues. 

\begin{figure}[t!]
  \centering
  \includegraphics[width=\linewidth]{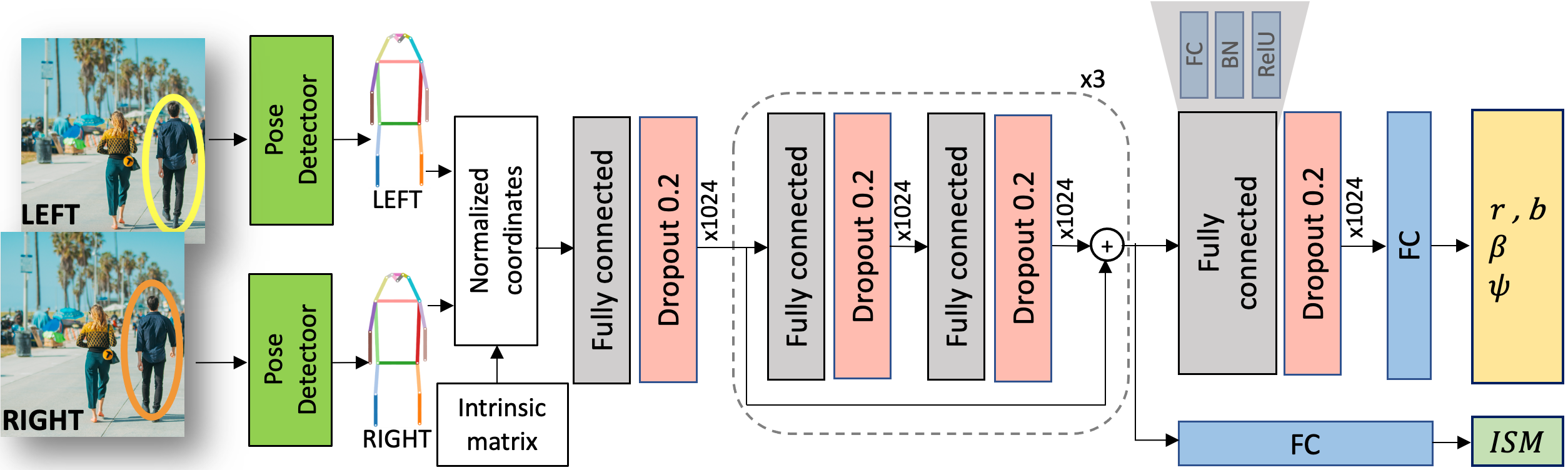}
  \caption{\small{Network architecture.  The input is a set of 2D keypoints extracted from a raw image. The outputs are the radial distance $r$ with its confidence interval $b$, the azimuthal angle $\beta$, the polar angle $\psi$, and the Instance-based stereo matching (ISM). Every fully connected layer is followed by a Batch Normalization layer (BN) \cite{ioffe2015batch} and a ReLU activation function.}}
  \label{fig:method}
\end{figure}

\textbf{Architecture. }
Our method consists of two steps. First, we reduce the input dimensionality by predicting 2D keypoints for each person in left-right images. Keypoints are a low-dimensional representation which is invariant to many nuisances, is suitable in the low-data regime and is prone to easy manipulations. Second, we analyze pairs of keypoints from left-right images in an ``all-vs-all'' setting to predict 3D location, and confidence interval of every person in the scene. 
Our simple architecture is shown in Figure \ref{fig:method} and consists of
few fully-connected layers with batch-normalization,  residual connections \cite{he2016residual}, and dropout \cite{srivastava2014dropout}.

\textbf{Input/output. } We use an off-the-shelf pose detector (\textit{e.g., }PifPaf \cite{kreiss2019pifpaf, kreiss2021openpifpaf}) to obtain a set of keypoints $\left[\vec{x}, \vec{y} \right]_i^T$ for every person $i$ in the left and right images. Each keypoint is projected into normalized image coordinates $ \left[\vec{x}^*, \vec{y}^*, \vec{1} \right]_i^T = I^{(i)}$ to prevent overfitting to a specific camera.
To construct the network inputs, we associate in an ``all-vs-all'' way the keypoints $I^{(l)}$ from each person $l$ in the left image with the one $I^{(r)}$ from each person $r$ in the right image, to form the associated pair $I^{(l,r)}$:
\begin{equation}
I^{(l,r)} =  I^{(l)} \; \| \; (I^{(l)} - I^{(r)})\;\; \forall \; l \in N_L, r \in N_R\;\;\;,
\end{equation}
where $\|$ is a concatenation operation, and $N_L$, $N_R$ denote the sets of detected instances in the left-right image pair.
If the sets of keypoints $I^{(l)}$ and $I^{(r)}$ belong to the same person, the input is a \textit{true pair} and we call it $I^{(l,r)}_{S-M}$, with subscript denoting stereo and monocular cues available. Otherwise the input is a \textit{false pair}, which we denote with $I^{(l,r)}_M$. We treat this problem as a binary classification task and use binary cross-entropy loss to train our network. We refer to this association task as  \textit{Instance-based stereo matching (ISM)} and to its loss as \textit{ISM loss}.
To disentangle the depth ambiguity from the other localization components (x, y), we use a spherical coordinate system  $(r, \beta, \psi)$, namely radial distance $r$, azimuthal angle $\beta$, and polar angle $\psi$. The size of an object projected onto the image plane depends on its radial distance $r$ and not on its depth $z$ \cite{monoloco}.

\begin{table*}[t!]
 \centering
  \begin{tabular}{|l|c c c c|c c c c|}
    \hline
    Method & \multicolumn{4}{c|}{ALE (m) $\downarrow$   \;\; [Recall (\%) $\uparrow$ ]} &  \multicolumn{4}{c|}{RALP-5\%  (\%) $\uparrow$ } \\
      & $Easy$ & $Mod.$  & $Hard$ & $All$  
      & $Easy$ & $Mod.$  & $Hard$ & $All$\\
    \hline\hline
    Monocular & & & & & & & &\\
    \hline
    Mono3D \cite{chen2016monocular} 
    & 2.26 [89\%] & 3.00 [65\%] & 3.98 [34\%]& 2.62 [69\%]
    & 9.21 & 1.26 & 0.21 & 7.22\\
    MonoPSR \cite{ku2019monopsr}
    &  0.89 [99\%] & 2.00 [93\%] & 2.40 [34\%] & 1.51 [83\%] 
    & 48.87 & 12.54 & 0.47 & 35.35 \\
    MonoLoco \cite{monoloco}
    &  0.83 [91\%] & 1.12 [72\%] & 1.15 [27\%] & 0.93 [70\%] & 49.01 & 19.44 & 1.89 &38.76 \\
    \hline
    Stereo & & & & & & & & \\
    \hline
    \textit{E2E-PL} \cite{e2e-pl} 
    &  \textit{0.12 [68\%]} & \textit{0.17 [23\%]} & \textit{0.60 [13\%]} & \textit{0.15 [43\%]}
    & \textit{49.32} & \textit{4.43}&	\textit{0.44} &	\textit{31.31} \\ 
    \textit{OC} \cite{oc-stereo}
    &  \textit{0.10 [66\%]} & \textit{0.14 [31\%]} & \textit{0.75 [6\%]} & \textit{0.13 [42\%]}
    & \textit{65.58} & \textit{26.38}&	\textit{1.46} &	\textit{41.30} \\ 
    3DOP \cite{chen20153dop}
    & 0.67 [88\%] & 1.19 [64\%] & 1.93 [37\%] & 0.93 [69\%] 
    & 57.88 & 22.70 & 3.85 & 45.92 \\
    PSF \cite{wenlongPSF}
    &  0.55 [88\%] & 0.65 [58\%] & 0.80 [25\%]& 0.56 [65\%] & 57.27 & 19.94 & 4.82 &46.15\\
    P-LiDAR \cite{wang2019pseudo}
    &\textbf{0.16} [88\%] & 0.72 [59\%]& 1.59 [33\%] &  0.46 [67\%] 
    &\textbf{88.94}& 42.91 & \textbf{10.41} & 66.33\\
    \hline
    B-ReID
    &  0.73 [91\%] & 0.78 [72\%] & 1.02 [28\%] & 0.77 [70\%]  & 73.81 & 39.44 & 4.48 & 58.23 \\
    B-Pose
    &  0.65 [91\%] & 0.77 [71\%]& 1.18 [27\%] &0.72 [70\%] & 73.92 & 39.10 & 4.82 & 58.25 \\
    B-Median
    &  0.57 [92\%] & 0.69 [72\%] & 0.78 [31\%] & 0.61 [72\%]  & 80.19 & 50.38 & 8.17 & 64.00 \\
    Our MonStereo
    &  0.29 [92\%]& \textbf{0.41} [70\%] & \textbf{0.50} [31\%]& \textbf{0.34} [71\%] 
    & 85.54 & \textbf{54.27}&	8.92 &	\textbf{67.60} \\
     \hline
  \end{tabular}

  \caption{\small{Comparing our proposed method against baselines on KITTI dataset ~\cite{Geiger2013Kitti}. 
  We use PifPaf \cite{kreiss2019pifpaf, kreiss2021openpifpaf} as off-the-shelf network to extract 2D poses. On the RALP metric, our MonStereo achieves state-of-the-art results. On the ALE metric, the confidence threshold of methods has been set to 0.5 and we show the recall between brackets to insure fair comparison. Italics entries are not directly comparable as they achieve a lower recall even when no threshold is set. Our method performs better on \textit{hard} instances while maintaining 2-5 times higher recall. The improvement of jointly solving the ISM and the 3D localization tasks is shown by the three baselines (B-). }}
  \label{tab:res_kitti}
\end{table*}

\textbf{Uncertainty. } We model the aleatoric uncertainty for the depth estimation task following MonoLoco \cite{monoloco} and using a relative Laplace loss based on the negative log-likelihood of a Laplace distribution:
$L_{\textrm{Laplace}}(x|r,b) = \frac{|1-r/x|}{b} + \log(2b)$,
where $x$ is the ground-truth and $(r,b)$ the predicted radial distance and the spread, respectively, making this training objective an attenuated $L_1$-type loss via spread $b$. 
% During training, the model has the freedom to predict a large spread $b$, leading to attenuated gradients for noisy data.
At inference time, the model predicts a radial distance $r$ and a spread $b$ 
which indicates its confidence about the predicted distance. The use of spherical coordinates allows to convey all the 3D localization uncertainty into the radial component $r$. 
% The spherical angles $\beta$ and $\psi$ can be derived from the projection of the object onto the image plane.

 \textbf{Inference. }
 The network performs 3D localization as well as ISM by predicting whether each pair of keypoints belongs to the same person or to different ones. The ISM component is also used to filter multiple results for the same person. At inference time, the network predicts $N_R$ outputs for each person in the left image (one for each associated pair) and selects the one with the highest predicted stereo matching. In fact, a \textit{true pair} $I^{(l,r)}_{S-M}$ always contains more information about the left instance than $I^{(l,r)}_M$. For a single image pair, the number of pairwise combinations grows quadratically as $N_L * N_R$ but, as the inputs are low-dimensional, the computation is parallelizable by including all the pairs in the same batch.

\textbf{Knowledge Injection. }
Monocular estimates are essential to address the long tail of stereo-based 3D localization, but they present their own issues. A typical dataset for 3D object detection such as KITTI \cite{Geiger2013Kitti} is not representative of the real world as it only contains few scenes recorded from a single city. For instance 
in the case of a child, any monocular estimate of depth will either fail
or rely solely on the ground plane estimation \cite{van2019neural}.
These settings make the network over-confident toward monocular estimates, as: (i) the predicted confidence intervals do not reflect the real distribution of human heights and the model can drastically fail in case of children or tall people; (ii) the training phase becomes ineffective as the network relies on monocular estimates even when a stereo association is available.
To tackle both issues, we inject knowledge in the training data by augmenting it with relevant examples from the long tail of the human height distribution.
We augment the KITTI dataset with synthetic 2D keypoints of people of heights ranging from 1.2 meters to 2 meters. We rely on the mild assumption that the aspect ratio between children and adults is unchanged and 
for each set of keypoints $I^{(l,r)}$, we sample a height $h$ from the uniform distribution $\mathcal{U}(1.2,2)$ and we derive a new ground-truth distance from the triangle similarity relation of human heights and distances. Then, we create a new input $I^{*(l,r)}$ updating the disparity and the ground-truth distance.  
We repeat this procedure for every pair $I^{(l,r)}$ with double-sided advantages. The network benefits from augmenting the number of \textit{true pairs} $I^{(l,r)}_{S-M}$ as it learns that disparity estimates correspond to correct depths whereas the monocular assumption of average height breaks down. It also benefits from augmenting {false pairs} $I^{(l,r)}_M$, by becoming receptive to more realistic human height variations, including children or very tall people. This knowledge is reflected in more calibrated confidence intervals of distance especially in the tail of human heights.

\textbf{MonStereo and MonoLoco. } Our MonStereo differs from MonoLoco~\cite{monoloco} by (i) the use of stereo images in combination with monocular information; (ii)
 the contribution of the ISM loss, an improved neural network architecture, and spherical coordinates to disentangle the ambiguity in the 3D localization task; (iii) a knowledge injection procedure to improve model robustness to outliers.

\section{Experiments}
  \begin{figure}[t!]
  \centering
  \includegraphics[width=\linewidth, height=5.3cm]{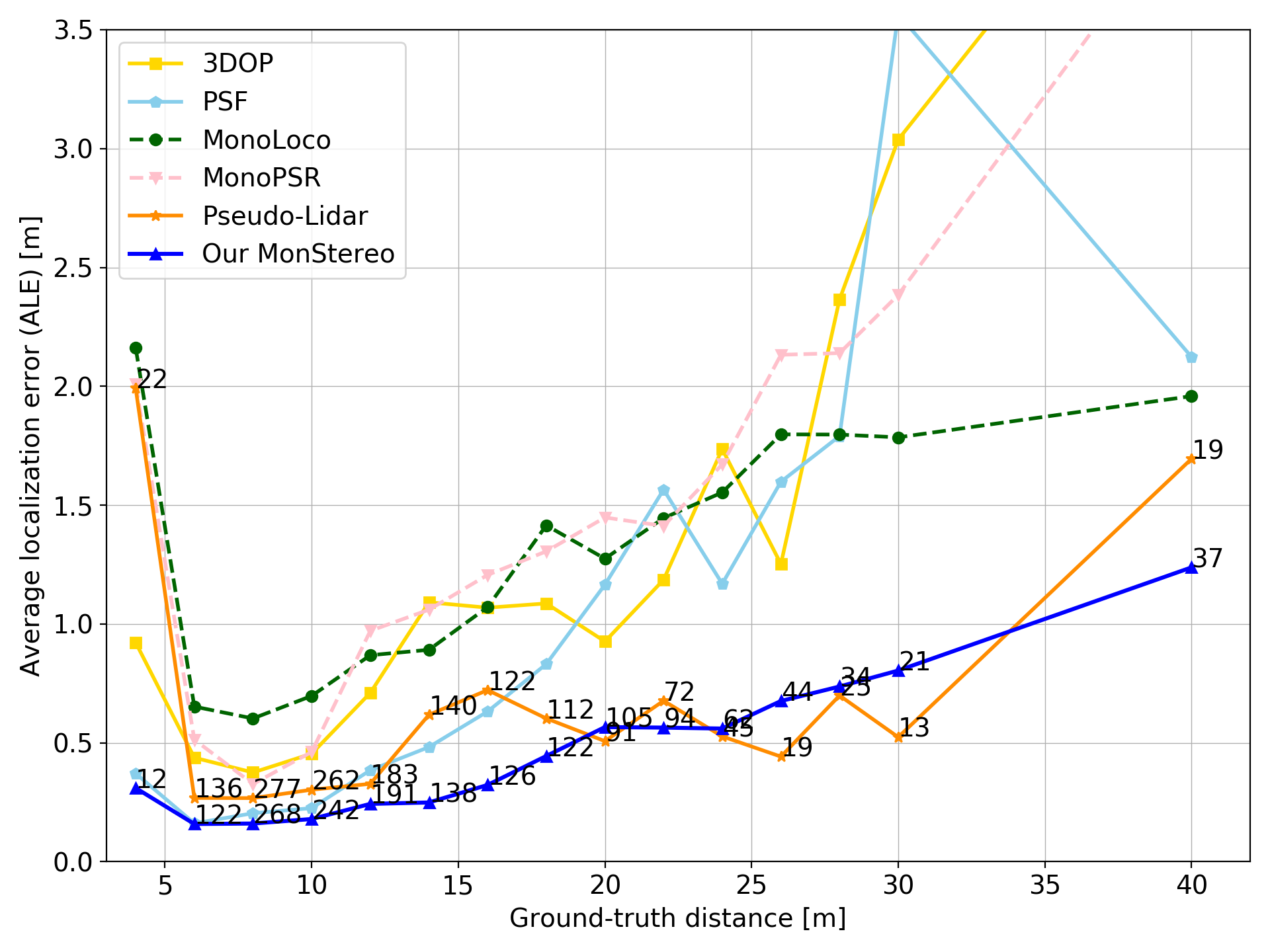}
\caption{\small{ALE as a function of distance. MonStereo achieves robust performance while even detecting more instances (numbers included) in the farthest clusters.}}%
\label{fig:results}
\end{figure}

\begin{figure}[t!]
\centering 
\includegraphics[width=\linewidth, height=5.3cm]{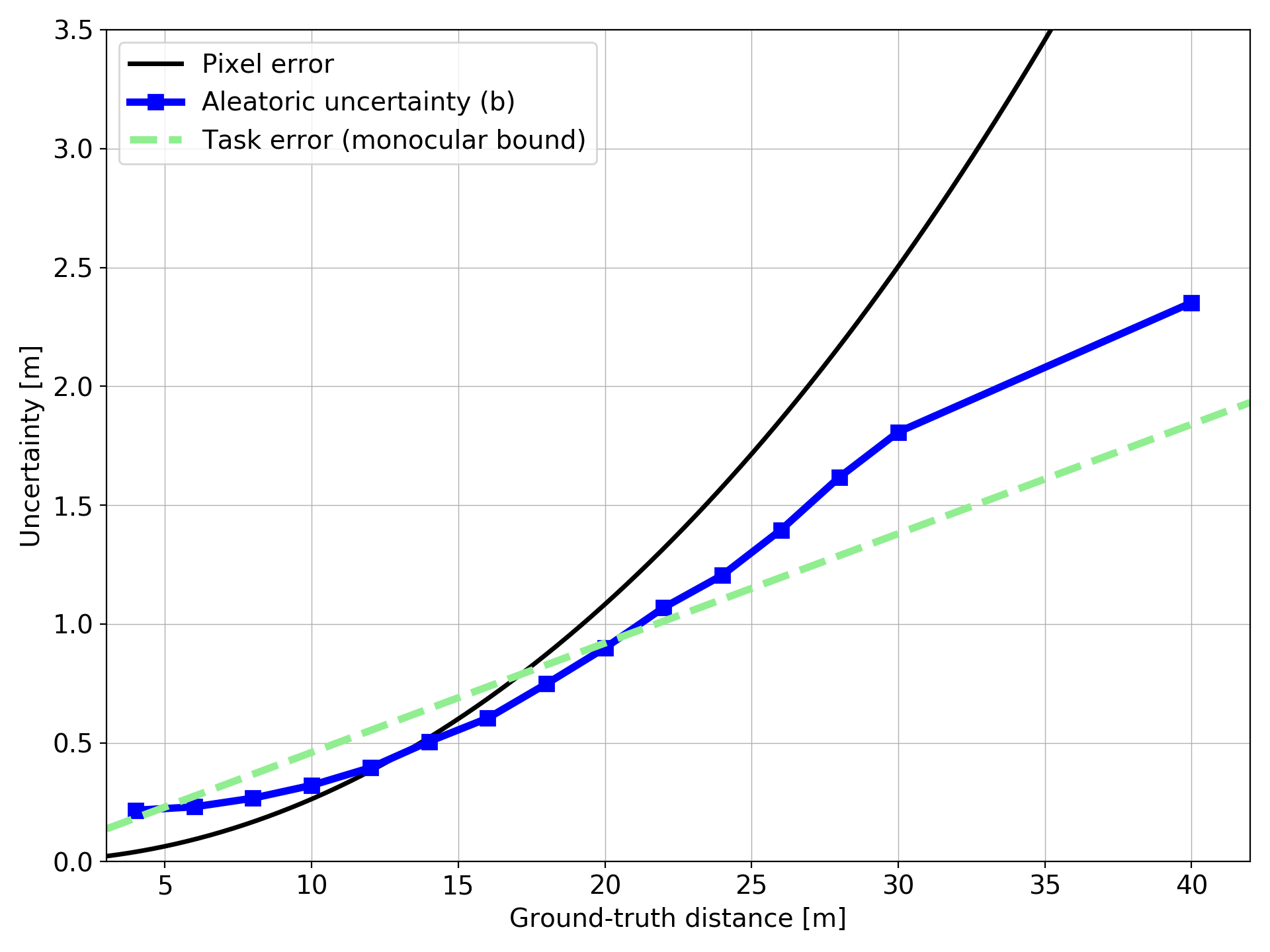}
 \caption{\small{For close instances the spread $b$ has a quadratical trend as MonStereo exploits stereo cues, and a linear trend at further distances thanks to monocular cues.}}
\label{fig:spread}
\end{figure}

Pedestrian 3D localization is a safety-critical task for self-driving cars and social robots, and we argue it is not sufficient to be accurate ``on average". In parallel to standard metrics, we evaluate the long-tail by analyzing box plots and predicted confidence intervals. In addition, we critically review the official KITTI 3D metrics for pedestrians and propose a practical 3D localization metric for pedestrians.

\subsection{Baselines}
\vspace{-3pt}
 \begin{figure*}[t!]
        \centering
        \begin{subfigure}[t!]{0.49\textwidth}
            \centering
            \includegraphics[height=5.2cm, width=\textwidth]{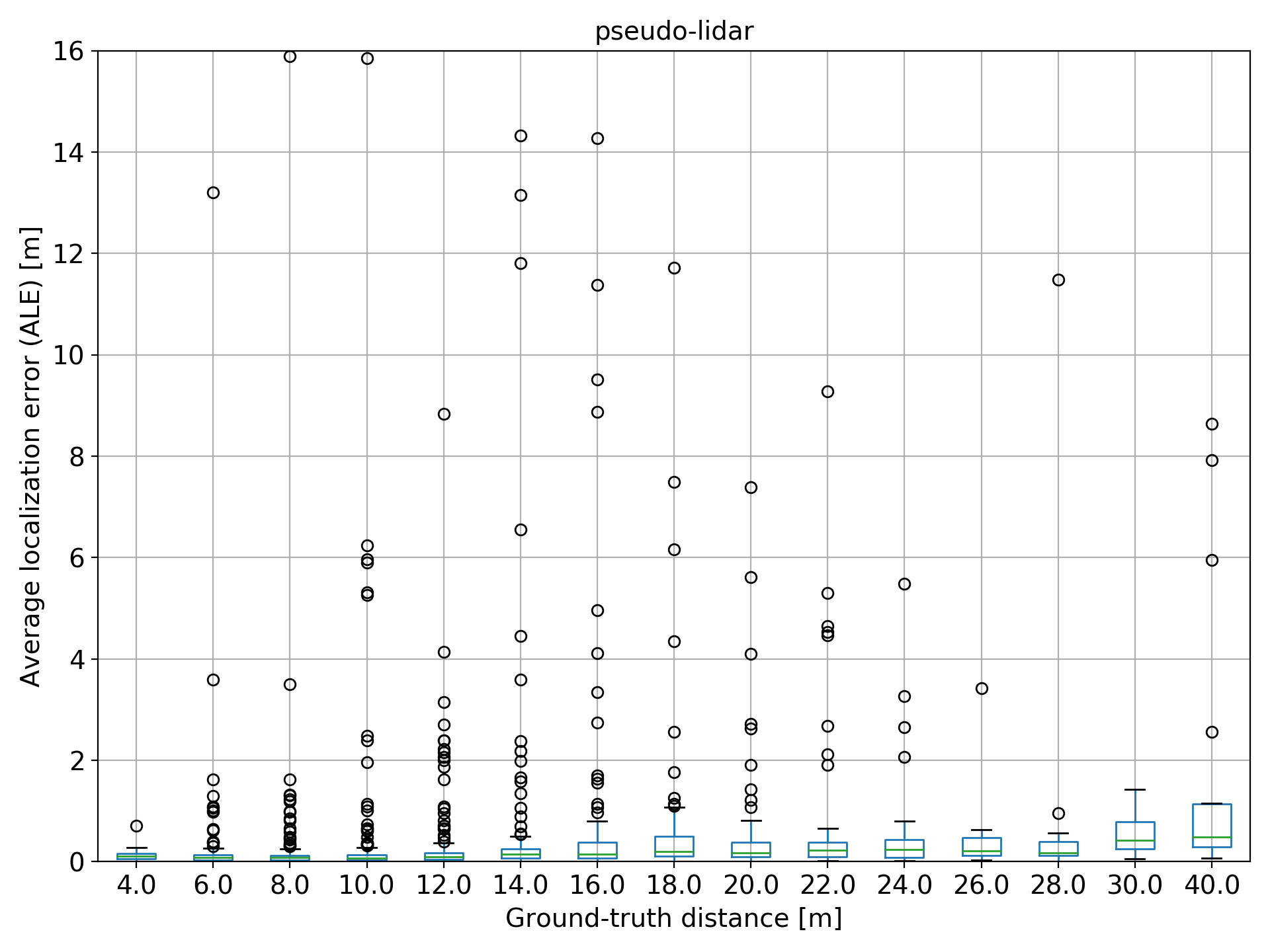}
        \end{subfigure}
        \begin{subfigure}[t!]{0.49\textwidth}  
            \centering 
            \includegraphics[height=5.2cm, width=\textwidth]{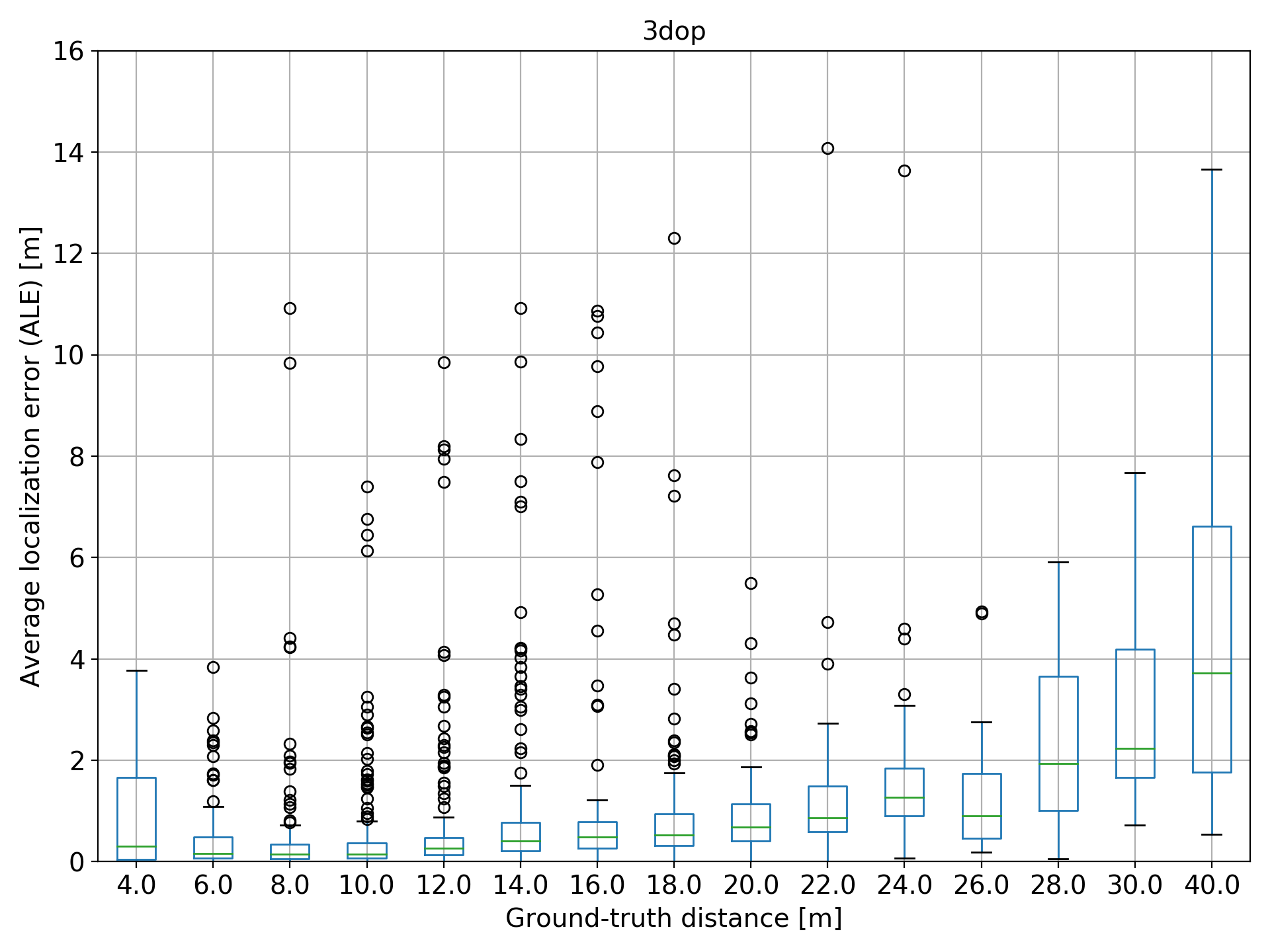}
        \end{subfigure}
        
        \vskip\baselineskip
        \vspace{-15pt}
        \begin{subfigure}[t!]{0.49\textwidth}   
            \centering 
            \includegraphics[height=5.2cm, width=\textwidth]{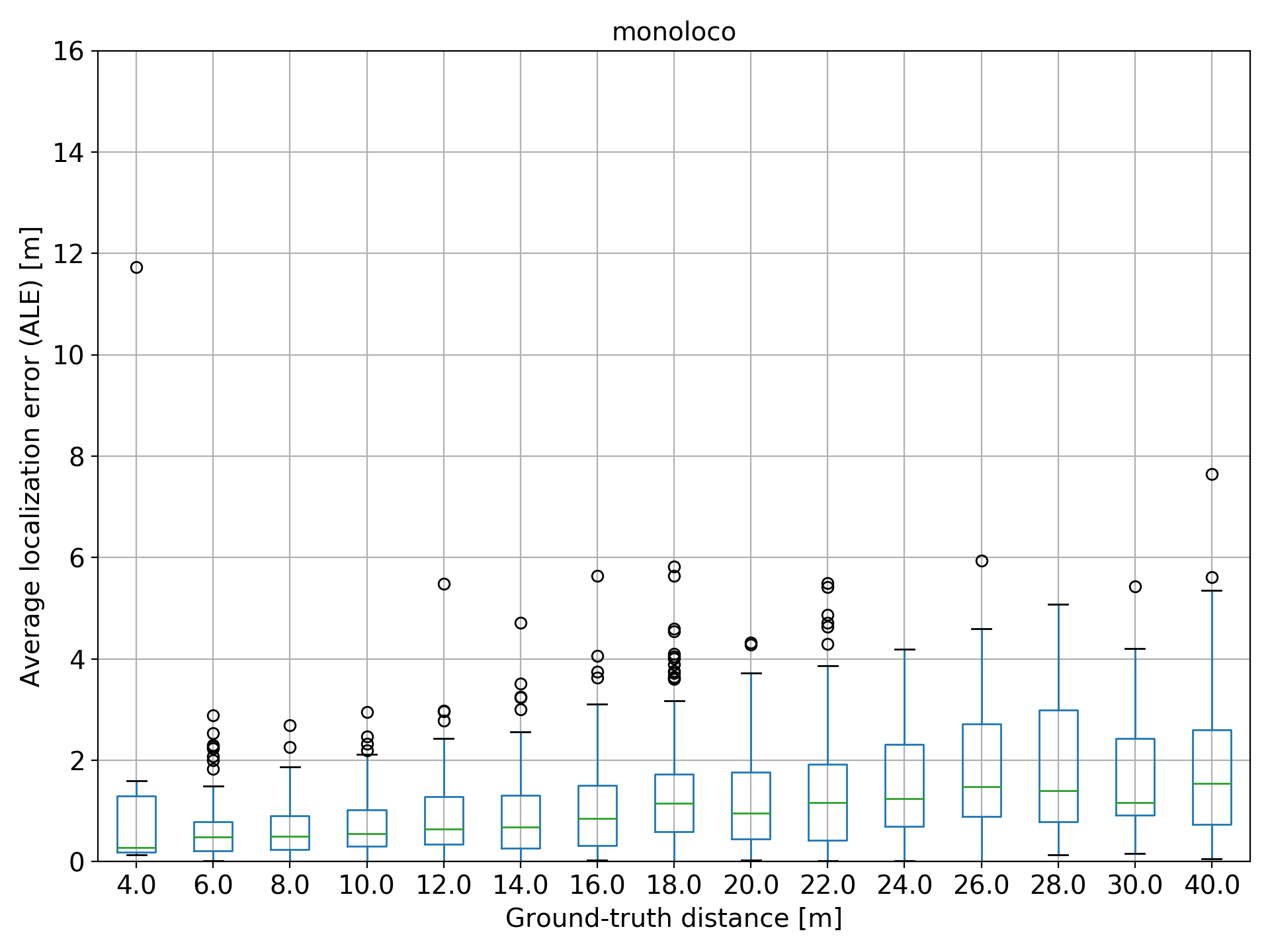}
        \end{subfigure}
        \begin{subfigure}[t!]{0.49\textwidth}
            \centering 
            \includegraphics[height=5.2cm, width=\textwidth]{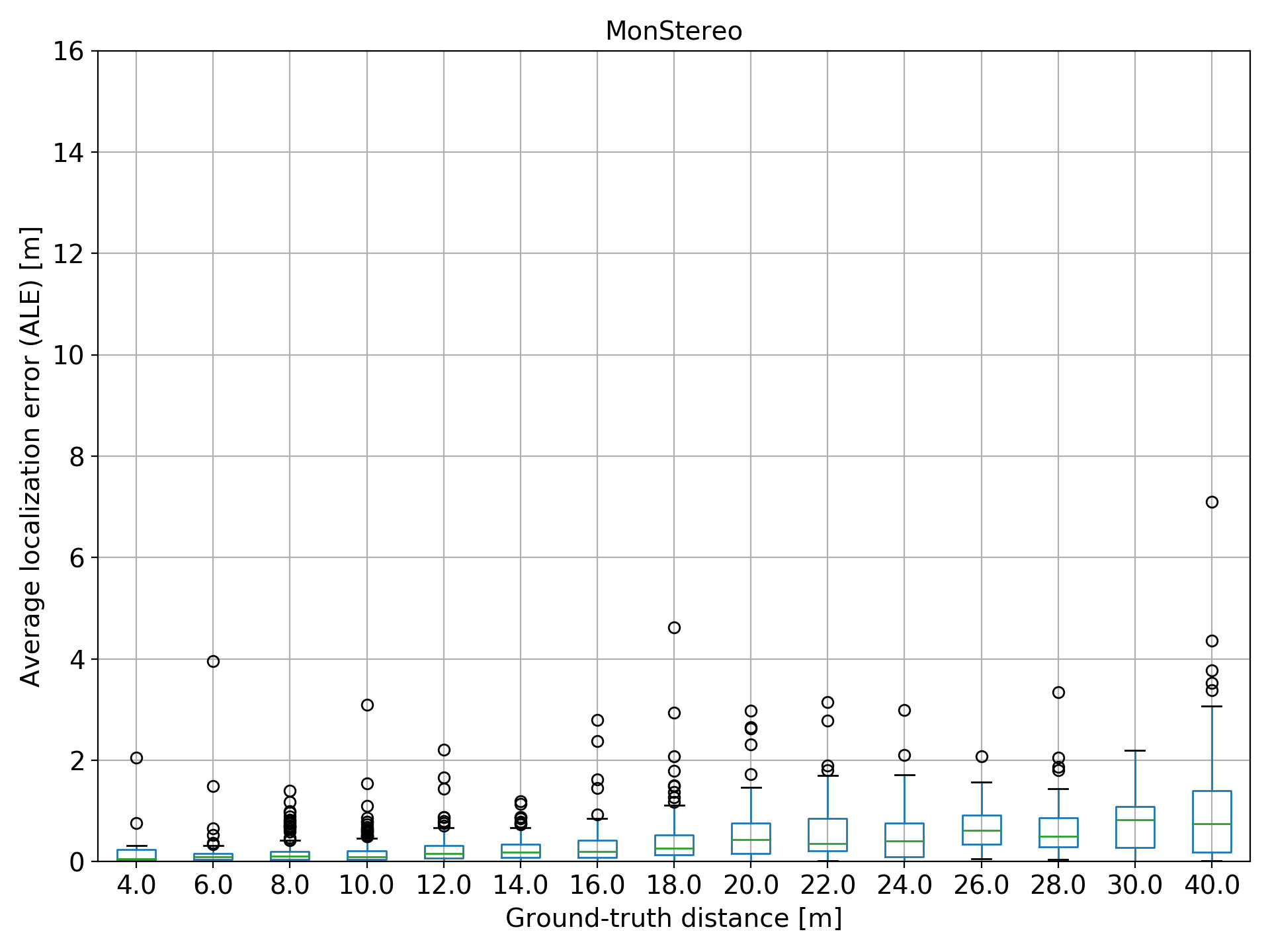}
        \end{subfigure}
        \caption{\small {Box plots of Average Localization Error (ALE). Circles identify outliers. Our MonStereo achieves very robust performance in the long tail with a maximum error of 7 meters for far instances and less than 5 meters in all the other cases. Every other stereo method has a few catastrophic estimates even for very close people. MonStereo's monocular component stabilizes the performances as shown by the performances of the monocular MonoLoco \cite{monoloco}, which is on average not as accurate as a stereo method but more robust.}}
        \label{fig:boxplot}
    \end{figure*}

Our learning framework jointly solves the instance-based stereo matching (ISM) and the 3D localization tasks in an end-to-end manner. As baselines, we analyze performances when solving these two tasks separately.

\textbf{ISM Baselines.} We develop two baselines to associate people in left-right images. Both methods provide the best similarity score for each person in the left image (reference) with respect to all the people in the right image (targets). 
\textit{B-Pose:} we use pose similarity based on the detected 2D keypoints, \textit{i.e.}, calculating how similar two poses are. We zero-center reference and target poses and we calculate the L2 norm between our reference vector and all the target vectors and save the scores.
\textit{B-ReID:} we associate the same person in left-right pairs of images by looking at the appearance of the person and the surrounding scene. We use a state-of-the-art Re-Identification model  \cite{adaimi2021deep} trained on Market-1501 \cite{zheng2015market} to make the association from cropped images.

\textbf{Median Baseline.} Our network not only solves the ISM task, but also estimates the depth from a set of keypoints. As a baseline (\textit{B-Median}), we apply our network and, if a match between two people is found, we calculate the depth from the median value of the set of disparities.

\textbf{Other Baselines.}
We also compare our method with several state-of-the-art monocular and stereo baselines in Table \ref{tab:res_kitti}, which provides results on the KITTI validation set.

\begin{table}[ht]
  \centering
   \resizebox{\columnwidth}{!}{%
  \begin{tabular}{|l|c c c c|}
    \hline
ALE [$\sigma$] (m) & $d < 10$ & $10<d<20$ & $20<d<30$ & $30<d<50$  \\ \hline
    S       & 0.24 [0.6] & 0.47 [0.9] & 1.38 [1.4] & 3.95 [2.4] \\
    S-x     & 0.52 [1.5] & 0.61 [1.4] & 1.72 [1.5] & 5.50 [3.0]\\
    S+M     & 0.25 [\textbf{0.4}] & 0.50 [\textbf{0.7}] & 1.08 [1.2]  & 2.24 [1.9]  \\
    MonStereo & \textbf{0.20} [\textbf{0.4}] & \textbf{0.38} [\textbf{0.7}] & \textbf{0.73} [\textbf{1.0}] & \textbf{1.63} [\textbf{1.8}] \\
    \hline
    
  \end{tabular}
}
\caption{\small{Impact of the ISM loss with mean and standard deviation of localization error. \textbf{S} simulates a standard stereo method by training a model solely with \textit{true pairs} $I^{(l,r)}_{S-M}$; the network could learn monocular cues but is not guided by the ISM loss. \textbf{S-x} is as S, but without providing y-coordinates of input keypoints to remove information on human heights. \textbf{S+M} is trained with the same set of pairs $I^{(l,r)}_{S-M}$ and $I^{(l,r)}_M$ of MonStereo without the ISM Loss. The long tail of far instances is the most impacted by the ISM loss.
    }}
  \label{tab:ism}
\end{table}

\subsection{Implementation Details}
\vspace{-3pt}
We train and evaluate our model on the KITTI Dataset ~\cite{Geiger2013Kitti} using the train/val split of Chen \textit{et al.} \cite{chen2016monocular}. To detect 2D keypoints, we use the off-the-shelf pose detector PifPaf \cite{kreiss2019pifpaf, kreiss2021openpifpaf} and we upscale the images by a factor of two to match the minimum dimension of 32 pixels for COCO instances. We train our network for 400 epochs using Adam optimizer~\cite{kingma2014adam}, a learning rate of $10^{-3}$, mini-batches of 512 and gradient clipping. We use a Laplace loss \cite{monoloco} for the radial distance, binary cross-entropy loss for stereo matching, and L1 loss for all other components. The losses are equally weighted. The KITTI dataset \cite{Geiger2013Kitti}  does not provide pairwise matching information, thus, we extend the ground-truth by associating each person in the left image with the corresponding one in the right image. We also perform horizontal flipping and switch left and right instances. 

\begin{table*}[t!]
  \centering
  \begin{tabular}{|l|c c c|c c c|c c c|c c c|}
    \hline

     & & ALE  & $\downarrow$ [m] &  & Recall  & $\uparrow$ [\%] & & I. Size  &  $\downarrow$ [\%]  \\
 & $Easy$ & $Mod.$ & $Hard$ & $Easy$ & $Mod.$ & $Hard$ & $Easy$ & $Mod.$ & $Hard$ \\
        \hline
    M
     & 0.77 & 0.82 & 1.35 & 58.0 & 58.9 & 32.5 & 4.6 & 5.0 & 4.9 \\
    W/o KI
     & 0.51 & 0.68 & 0.87 & 76.2 & 72.7 & 42.0 & 4.1 & 4.5 & 4.4 \\
   With KI
    & \textbf{0.29} & \textbf{0.41} & \textbf{0.50} & \textbf{91.2} & \textbf{81.9} & \textbf{65.4} & \textbf{3.8} & \textbf{4.1} & \textbf{4.1}\\
    \hline
  \end{tabular}
    \vspace{2pt}
    \caption{\small{Impact of knowledge injection (KI). We trained a monocular baseline M and a stereo one without KI. Recall measures the fraction of instances inside the intervals, and \textit{I. Size} is the ratio between the spread \textit{b} and the ground-truth distance. KI improves performances, especially for \textit{Hard} instances. The intervals do not grow, as the spread $b$ is reduced by better exploiting stereo cues.}}
  \label{tab:ki}
\end{table*}

 \subsection{3D Localization Metric for Pedestrians}
 \vspace{-3pt}
The majority of previous works for vision-based 3D object detection only reports results on the car category \cite{roddick2018orthographic,li2019stereo}. We argue that KITTI official metrics, \textit{i.e.}, bird's eye view and 3D average precision  \cite{Geiger2013Kitti}, are not appropriate for pedestrians, as a pedestrian 3D bounding box has an average width and length of 60 cm and 75 cm. Considering perfect orientation and dimensions, a distance error of 18 cm already leads to an intersection over union lower than 0.5. This requirement is unnecessarily strict and shifts the attention of the community from the challenging instances to the easy ones, where obtaining results with a precision of few centimeters may still be possible. 
Furthermore, the KITTI official metric assigns to each instance a difficulty regime based on bounding box height, level of occlusion and truncation: \textit{easy}, \textit{moderate} and \textit{hard}. Each category includes instances from the simpler categories,
and, due to the predominant number of easy instances (1240 ``easy" pedestrians and 300 ``hard" ones), the metric can underestimate the impact of challenging instances. 
To address the current limitations, we propose to consider a safety-critical area around a pedestrian, recognizing a prediction as correct if the localization error between the predicted and ground-truth box is less than a threshold error. Differently from the metric proposed by Xiang \textit{et al.} \cite{Xiang2015Datadriven3V}, we define an adaptive threshold based on distance. In our evaluation, we use a relative error $e_z$ of 5\%, considering 1 meter as reasonable safety-critical range for people 20 meters far, but thresholds are application dependent. We refer to the metric as Relative Average Localization Precision (RALP) and we split the evaluation into ``Easy", ``Moderate", ``Hard", with no overlap between categories,  and ``All".
At last, we evaluate the Average Localization Error (ALE) \cite{monoloco}, that differently from average precision metrics, penalizes large errors and is suited for the long tail of 3D localization.

\subsection{Results}
\vspace{-3pt}
Table \ref{tab:res_kitti} summarizes our 3D localization results with the ALE and RALP metrics.
Our method outperforms every other stereo method in the ALE metric for \textit{Moderate}, \textit{Hard}  and \textit{All} instances. Solving jointly the ISM task and the 3D localization one is a crucial ingredient, as shown by the three baselines where the association and localization tasks are sequential.
We make in-depth comparisons with the ALE metric as a function of the ground-truth distance in Figure \ref{fig:results}. On the ISM task, we obtain an accuracy of 98.2\%. 

\textbf{Outliers. } To go beyond ``average-based metrics", we analyze the entire distribution of predictions through the box plots in Figure \ref{fig:boxplot}. Our MonStereo is drastically more reliable for the long tail of predictions, especially when compared to other stereo methods. MonStereo's maximum error is lower than 5 meters, while Pseudo-LiDAR \cite{wang2019pseudo} and 3DOP \cite{chen20153dop} have maximum errors of 24 and 17 meters respectively.

\textbf{Long Tail-aware Confidence Intervals. }
The spread $b$ is the result of a probabilistic interpretation of the model, and it is calibrated during training according to the Laplace distribution. Training data includes the long tail of the height distribution, and the number of \textit{true} and \textit{false} pairs is balanced by design. During validation, stereo cues are available for the majority of instances, and the confidence intervals become more conservative, with $ 86.0\%$ of validation instances lying inside them. Yet the length of each side is only 3.9\% of the predicted distance, making the confidence intervals small enough for practical purposes.

\textbf{Monocular and Stereo Limitations. }
We compare in Figure \ref{fig:spread} the predicted aleatoric uncertainty $b$ with the 3D localization ambiguity for monocular and stereo modalities through the monocular \textit{task error} and the stereo \textit{pixel error}, respectively. The stereo ambiguity is very small at close distances but grows quadratically, while the monocular one grows linearly. Our MonStereo intrinsically learns to predict 3D locations of instances combining stereo and monocular cues based on the distance. This is reflected in the estimated confidence intervals. For close instances the trend is quadratic as stereo cues are more accurate. In contrast, at further distances the trend is linear. For very far pedestrians the predicted $b$ is larger than the \textit{task error} as, in addition to the task-based uncertainty, the aleatoric uncertainty $b$ also includes input noise \cite{Kendall2017WhatUD}, in our case 2D keypoints noise.

\textbf{Run Time. }
We conducted our experiments using a machine with a single NVIDIA GeForce GTX 1080 Ti and Intel(R) Core(TM) i7-8700 CPU @ 3.20GH for both 2D pose detector and MonStereo. Our run time relies heavily on the 2D pose detector \cite{kreiss2019pifpaf}  ($\sim$ 150 ms) with negligible computation from 2D to 3D, making our pipeline suitable for real-time applications. The runtime is on average 1.5 ms, and can grow up to to 16 milliseconds in the most crowded case of 28/29 people in each of the two images.

\subsection{Ablation Studies}
\vspace{-3pt}
Learning an ensemble of monocular and stereo cues is a delicate balance. The ISM loss prevents our method from overfitting to stereo disparity, and the KI prevents it from overfitting to monocular cues.

\textbf{ISM Loss. }
This loss encourages the use of monocular cues when stereo ones are not available or monocular cues are more convenient \textit{(e.g.}, faraway people where pixel disparity is not accurate enough).  Without explicit guidance, the network over-relies on stereo cues, as shown in Table \ref{tab:ism}.

\textbf{Knowledge Injection (KI). }
Without KI, the network over-relies on monocular cues. We illustrate it by training a monocular baseline, and a stereo baseline without KI. We analyze ALE metric, recall (percentage of instances inside the intervals) and relative size of the intervals in Table \ref{tab:ki}. KI improves results and calibrates the confidence intervals, including the long tail of the height distribution. Recall increases, yet the interval size decreases, as KI helps to exploit stereo cues and reduce the spread \textit{b}. 

\section{Conclusions}
\vspace{-3pt}
We have proposed a vision-based approach tailored for the long tail of 3D human localization. 
Our neural network implicitly learns to leverage monocular and/or stereo cues, while communicating uncertainty driven by those cues. Our method goes beyond providing competitive results ``on average'' and shows reasonable estimates even for outliers. We hope to direct the attention of the vision community towards the long tail for autonomous driving and social robot applications, still an unchartered research territory. 

\textbf{Acknowledgements.}
This work was supported by the Swiss National Science Foundation under the Grant 2OOO21-L92326, the SNSF Spark fund 190677, and Valeo. 

\bibliographystyle{IEEEtran}
\bibliography{references}
\end{document}